\documentclass[3p,times,procedia]{elsarticle}
\usepackage{amsmath}
\usepackage{graphicx} % Graphics
\usepackage{xspace} % for xspaces in new commands
\usepackage{todonotes}
\usepackage{subfig} % For using subfigures and alternative to subcaption if there are template issues

\newboolean{reduced_version}
\ifthenelse{\isundefined{\makeReduced}}{% Configure \makeReduced with command from Makefile
	\setboolean{reduced_version}{false}}{
	\setboolean{reduced_version}{true}}
\newboolean{with_rebuttal}
\ifthenelse{\isundefined{\makeRebuttal}}{
	\setboolean{with_rebuttal}{false}}{
	\setboolean{with_rebuttal}{true}}
\newboolean{condensed_image_version}
\ifthenelse{\isundefined{\makeCondensedImages}}{% Configure \makeCondensedImages with command from Makefile
	\setboolean{condensed_image_version}{false}}{
	\setboolean{condensed_image_version}{true}}
\newboolean{condensed_table_version}
\setboolean{condensed_table_version}{false}
\newboolean{wide_tables}
\ifthenelse{\isundefined{\makeWideTables}}{% Configure \makeWideTables with command from Makefile
	\setboolean{wide_tables}{false}}{
	\setboolean{wide_tables}{true}}
\ifthenelse{\isundefined{\setImageScaleFactor}}{% unused as of 2018/03/20
	}{
	}

\newcommand{\termSV}[0]{Shapley value\xspace}
\newcommand{\termSVs}[0]{Shapley values\xspace}
\newcommand{\termSVbased}[0]{Shapley-value-based\xspace}

\newcommand{\TermSV}[0]{\termSV} % force first letter uppercase
\newcommand{\TermSVs}[0]{\termSVs} % force first letter uppercase

\newcommand{\pow}[1]{\mathcal{P}(#1)}
\usepackage{amssymb}
\newcommand{\RR}{\mathbb{R}}

\usepackage{ecrc}
\usepackage[bookmarks=false]{hyperref}
    \hypersetup{colorlinks,
      linkcolor=blue,
      citecolor=blue,
      urlcolor=blue}
\volume{00}

\firstpage{1}

\journalname{Procedia Computer Science}

\runauth{Stier et al.}

\jid{procs}

\usepackage{amssymb}
\usepackage[figuresright]{rotating}
\begin{document}
\begin{frontmatter}



\dochead{22nd International Conference on Knowledge-Based and\\
Intelligent Information \& Engineering Systems}%
%% e.g. \dochead{17th International Conference on Dynamical Processes in Excited States of Solids}

\title{Analysing Neural Network Topologies:\protect\\ a Game Theoretic Approach}


\author[a]{Julian Stier\corref{cor1}} 
\author[b,c]{Gabriele Gianini}
\author[a]{Michael Granitzer}
\author[a]{Konstantin Ziegler}

\address[a]{University of Passau, Innstrasse 42, 94032 Passau, Germany}
\address[b]{Universit\`{a} degli Studi di Milano, via Bramante 65, Crema, IT26013, Italy} 
\address[c]{EBTIC, Khalifa University of Science, Technology \& Research, Hadbat Al Zaafran 127788, Abu Dhabi (UAE)}

\begin{abstract}
Artificial Neural Networks have shown impressive success in very different application cases.
Choosing a proper network architecture is a critical decision for a network's success, usually done in a manual manner.
As a straightforward strategy, large, mostly fully connected architectures are selected, thereby relying on a good optimization strategy to find proper weights while at the same time avoiding overfitting.
However, large parts of the final network are redundant.
In the best case, large parts of the network become simply irrelevant for later inferencing.
In the worst case, highly parameterized architectures hinder proper optimization and allow the easy creation of adverserial examples fooling the network.

A first step in removing irrelevant architectural parts lies in identifying those parts, which requires measuring the contribution of individual components such as neurons.
In previous work, heuristics based on using the weight distribution of a neuron as contribution measure have shown some success, but do not provide a proper theoretical understanding.

Therefore, in our work we investigate game theoretic measures, namely the \termSV (SV), in order to separate relevant from irrelevant parts of an artificial neural network.
We begin by designing a coalitional game for an artificial neural network, where neurons form coalitions and the average contributions of neurons to coalitions yield to the \termSV.
In order to measure how well the \termSV measures the contribution of individual neurons, we remove low-contributing neurons and measure its impact on the network performance.

In our experiments we show that the \termSV outperforms other heuristics for measuring the contribution of neurons. 
\end{abstract}




\cortext[cor1]{Corresponding author}
\end{frontmatter}

%\correspondingauthor[*]{Corresponding author. Tel.: +0-000-000-0000 ; fax: +0-000-000-0000.}
\email{julian.stier@uni-passau.de}

% \linenumbers


%\enlargethispage{-7mm}

% Condensed tables based on config
\ifthenelse{\boolean{condensed_table_version}}{
	\newcommand{\tableType}[0]{table}
}{
	\newcommand{\tableType}[0]{table*}
}
\ifthenelse{\boolean{reduced_version}}{
	\newcommand{\reducibleTableType}[0]{table}
}{
	\newcommand{\reducibleTableType}[0]{table*}
}
\ifthenelse{\boolean{condensed_image_version}}{
	\newcommand{\figureType}[0]{figure}
}{
	\newcommand{\figureType}[0]{figure*}
}

% Intro
\section{Introduction}\label{sec:introduction}
The architecture of an Artificial Neural Network (ANN) strongly influences its performance \cite{lecun1995convolutional,hochreiter1997long,srivastava2015highway}.
However, designing the structure of an artificial neural network is a complex task requiring expert knowledge and extensive experimentation.
Usually fully connected layers are used yielding to a high number of parameters.
While well-chosen optimization strategies allow to identify proper parameterization in general, larger architectures (i) have an increased risk of overfitting the training data, (ii) use more computational resources and (iii) are more affected by adversarial examples.\\

Identifying optimal architectures for an ANN is a NP-complete optimization problem.
Solutions can be categorized into bottom-up, top-down or mixed methods.
Bottom-up methods start from small architectures or none at all and gradually add more components (e.g. layers, neurons or weights).
An example for a bottom-up method is a grid-search \cite{bergstra2012random} for probing different number of hidden layers and number of neurons per layer.
Top-down methods start with larger architectures and remove low-contributing components, yielding a significantly smaller, \textit{pruned} architecture.

Well known top-down methods such as \textit{optimal brain damage} \cite{lecun1989optimal} or \textit{skeletonization} \cite{mozer1989skeletonization} utilize different heuristics, like for example the weight of a connection, and different search strategies, like for example greedy search, in order to identify components to remove.
Both methods, top-down and bottom-up, conduct a non-exhaustive search in a huge parameter space.
Such a search requires proper importance measures, i.e. measures to identify the importance of individual components.
However, most importance-measures do not rely on a well-formed theory but are defined in an ad-hoc manner, thereby limiting its applicability.

When viewing neurons in a neural network as competing and collaborating individuals, Game Theory can provide a possible theoretical background for properly selecting the most important ``player'' in a game.
In particular, coalitional games (also known as cooperative games) allow to view groups of neurons as coallitions competing with other coalitions.
Measures like for example the \termSV \cite{shapley1953} allow to determine the payoff for an individual, thereby determining its contribution to the coalition.

\subsection{Contributions}\label{sec:contributions}
In our work we utilize the \termSV as importance-measure to determine the contribution of neurons in a network.
We transform an ANN into a coalitional game between neurons, giving us access to well-studied game theoretic results.
The possibilities of this transformation process are discussed the first time, despite former existing experiments involving the \termSV, and treating the ANN as a coalitional game is separated from pruning ANNs.

Given the coalitional game from this transformation, we can estimate the \termSV for every neuron reflecting its individual contribution to the overall ANN architecture.
As the \termSV requires forming all possible coalitions, we suggest a sampling procedure for obtaining \termSV approximations.
The suggested sampling parameters are justified, even for non-uniform \termSV distributions.

Finally, we use the \termSV in a top-down pruning strategy and compare it to other heuristical pruning measures based on the weights between neurons.
We show that the \termSV provides a more robust estimate for the importance of a neuron yielding to a better performance of an ANN for the same model size than weight-based heuristics.
While pruning with \termSVs was previosuly only shown in small problem domains \cite{Leon14} and with one single strategy based on \termSVs, this work presents results on image and text classification with multiple strategies, including \termSVs.

\ifthenelse{\boolean{reduced_version}}{
	% No structure in reduced version given
}{
	\subsection{Structure}\label{sec:structure}
	% Structure
	Section \ref{sec:shapley-anns} describes the design of coalitional games on artificial neural networks.
	Different design choices are discussed and some empirical recommendations given.
	The design of a coalitional game on an ANN provides Shapley values for single structural components.

	Experiments in section \ref{sec:experiments} show that \termSVs can be obtained with Monte Carlo methods with sufficient precision, even if the result of a single inference step of an ANN might be computationally expensive to obtain.

	Based on the designed game, top-down methods, called \textit{pruning strategies}, with Shapley values are presented and compared to other top-down methods in section \ref{sec:experiments}.
	This section also contains the evaluation of properties of the strategies and the strategy comparisons.
	Finally, future work and a conclusion are given in sections \ref{sec:future-work} and \ref{sec:conclusion}.\\
}

% Related work
\section{Related work}\label{sec:related-work}
Bergstra et.\@ al claim ``grid search and manual search are the most widely used strategies for hyper-parameter optimization'' \cite{bergstra2012random}.
However, there exist various destructive and constructive approaches to obtain better network topologies.

Concerning destructive approaches, \textit{optimal brain damage} \cite{lecun1989optimal} ``uses information-theoretic ideas to derive [..] nearly optimal schemes for adapting the size of a[n artificial] neural network''.
For this, it uses second-derivative information and tries to minimize a composed cost function of training error and a measure of network complexity.

In \textit{Skeletonization} \cite{mozer1989skeletonization}, Mozer presents a destructive technique in which he prunes network components by means of their relevance.
The relevance is basically measured as the difference of the training error with and without the specific network component.
With respect to pruning strategies in section \ref{sec:experiments}, this technique is similar to the payoff function used in obtaining \termSVs.

Another second-order derivative method is presented with \textit{optimal brain surgeon} \cite{hassibi1993optimal} and improves the previous optimal brain damage method by pruning multiple weights based on their error change and immediately applying changes to remaining weights.

In \cite{keinan2004causal,keinan2004fair} Keinan et. al present the multi-perturbation \termSV analysis (MSA) using \termSV with ``a data set of multi-lesions or other perturbations''.
\TermSV is used to analyse contributions of components in biological neural networks.
Different choices of network components as aspect of the coalitional game are considered in \cite{keinan2006axiomatic}, chapter 9.

Based on the multi-perturbation \termSV analysis, Cohen et. al \cite{cohen2007feature} present a Contribution-Selection algorithm (CSA) ``using either forward selection or backward elimination'' \cite{cohen2007feature}.
Furthermore, K\"{o}tter et. al use the ``Shapley value principle [..] to assess the contributions of individual brain structures'' \cite{kotter2007shapley}.
They even find ``strong correlation between Shapley values'' and properties from graph theory such as ``betweenness centrality and connection density'' \cite{kotter2007shapley}.

Leon \cite{Leon14} uses \termSV to optimize artificial neural network topologies by pruning neurons with minimal value or below a threshold in relation to the average value of the \termSV distribution.
The method is applied on the XOR-, Iris-, energy efficiency, ionosphere, and Yacht hydrodynamics problems which all yield test set accuracies above 0.9 with at most four neurons in their networks' hidden layer.

Schuster and Yamaguchi \cite{schuster2010application} investigate a complementary approach, where the interaction of two neurons in an artificial neural network is seen as a non-cooperative game.

\section{A Game on Topologies}\label{sec:shapley-anns}
In a coalitional game, different subsets of a population of players generate different payoffs.
The payoff for a subset (\emph{coalition}) depends only on the participating players and a central question is how to value a single players ``contribution''.

\subsection{The Shapley Value}
% Formal introduction
The predominant solution concept for coalitional games is the \textit{\termSV} \cite{shapley1953}.
Let $U$ be a set of $n$ players, $\pow{U}$ its powerset, and let $v \colon \pow{U} \to \RR$ be a set function which assigns a payoff $v(S)$ to every subset $S \subseteq U$ of players.
The \termSV ``can be interpreted as the expected marginal contribution of player $i$'' \cite{roth1988}.
For player $i \in U$, it is given by
\begin{align}
	\label{eq:shapley-value}\phi_v(i) & =
		\frac{1}{n!}\sum_{S\subseteq U\setminus\{i\}}\left(|S|!(n-|S|-1)!\right) \cdot (v(S)-v(S-i)) \\
	\label{eq:shapley-value-alternative} & =
		\frac{1}{|U|!}\sum_{\pi\in\Pi}\left(v(P^{\pi}_i\cup\{i\})-v(P^{\pi}_i)\right),
\end{align}
where $\Pi$ is the set of all permutations of $U$ and $P^{\pi}_i = \{j\in U: \pi(j) < \pi(i)\}$.

% Example calculation for Shapley Value to meet requirement after review:
% ``Page 3, Section 3.1, Equations 1 through 4: The Authors are asked to provide a detailed explanation of the meaning of the aforementioned equations. The way the equations are presented is not clear to the unfamiliar reader.''
Imagine a \textbf{simple example} with three collaborating players $U = \{A, B, C\}$ contributing to a common goal such as selling a product.
The payoff of players is only known coalition-wise and thus the payoff function $v(S)$ could be given as:
$\{(\emptyset, 0), (\{A\}, 1), (\{B\}, 2), (\{C\}, 2), (\{A, B\}, 4), (\{A, C\}, 3), (\{B, C\}, 3), (\{A, B, C\}, 5)\}$.
Then the \termSV for each player results in $\phi_v(A) = 1.5$, $\phi_v(B) = 2$ and $\phi_v(C) = 1.5$ which can be scaled to $\phi_v(A) = 0.3$, $\phi_v(B) = 0.4$ and $\phi_v(C) = 0.3$, given the fact that the maximum payoff is $5$.
Player $B$ can then be interpreted as most contributing player to the game and both players $A$ and $C$ are contributing less.

Due to its exponential computational complexity, \termSVs are approximated with a Monte Carlo method.
The \textit{subset definition}~\eqref{eq:shapley-value} for $\phi_v(i)$ is approximated with random subsets $R\subset U$:
\begin{equation}
	\label{eq:shapley-value-approximated}\phi^R_v(i) =
		\frac{1}{\sum_{S\in R} \omega_S} \sum_{S\in R}\omega_S\cdot(v(S)-v(S-i)),
\end{equation}
with $\omega_S = |S|!(n-|S|-1)!$.
Analogously, the \textit{permutation definition}~\eqref{eq:shapley-value-alternative} is approximated with $r$ random permutations $\Pi^R$:
\begin{equation}
	\label{eq:shapley-value-alternative-approximated}\phi^{\Pi^R}_v(i) =
		\frac{1}{r}\sum_{\pi\in\Pi^R}\left(v(P^{\pi}_i\cup\{i\})-v(P^{\pi}_i)\right).
\end{equation}

\subsection{Designing the Game}\label{sec:design}
The idea of assigning each player a value, given a set function which defines the payoff of a coalition of players, can be transferred to neural networks.
For this, a set of players $U$ and the payoff function $v$ must be defined.

The \textbf{set of players $U$} can consist of any mutually excluding structural components of the network.
Choosing the concrete structural components defines the \textit{perspective} in which the game is played.
Because the structure can be arbitrarily broken into players, perspectives are categorized into homogeneous and non-homogeneous perspectives.

Homogeneous perspectives consist exclusively of structurally equivalent components.
Non-homogeneous perspectives are not further considered in this paper as they are not directly intuitive and introduce unnecessary complexity.
An example for a homogeneous perspective is the set of players representing each neuron in the hidden layer of a feed-forward network with one hidden layer.
Neurons of the input layer or several layers of a multilayer neural network provide other perspectives.

To define the \textbf{payoff function $v$} of the coalitional game of an ANN, any evaluation measure or error value of the network could be considered.
Error values such as the training error are unbounded which might be undesirable for later analysis.
Evaluation measures such as the accuracy are usually bounded and the cross-entropy accuracy in particular is used in this work.
Usually, coalitional games in game theory can be combined based on their super-additive payoff functions.
However, both choices -- error values or evaluation measures -- do not provide the super-additivity property.
This disables deriving desirable properties of symmetry, efficiency and additivity for the \termSV as proven by Shapley for games with such super-additive payoffs \cite{shapley1953}.
It is not possible to combine multiple coalitional games on artificial neural networks without finding a super-additive payoff function.

The accuracy\footnote{Number of correctly classified instances given a test or validation set of e.g. 10000 observations.} as an evaluation measure of ANNs is used to construct the payoff value of the coalitional game.
Looking at the accuracy it can be stated:
\begin{enumerate}
	\item The payoff for the \textit{grand coalition} is not necessarily the maximum possible payoff value.
	\item The maximum value of the payoff is not known in an analytical way prior to computing all values for every possible coalition.
	\item The accuracy is not super-additive, meaning there are coalitions $S, T \subseteq U$ with $v(S) < v(S\cap T) + v(S - T)$.
		It is not even monotone as there might exist neural networks with fewer network components but still larger accuracy.
\end{enumerate}

Nonetheless, given a network evaluation measure $m$ such as the accuracy, a payoff value can be defined as following:
\[
	v(S) := m(S) - m(\emptyset),
\]
with $S \subseteq U$ and $m(T)$ denoting the evaluated measure of the network with only players contained in $T$.
Usually, $m(\emptyset)$ should be at least above the na\"ive expectation of the classification or regression problem.
Therefore, it can be assumed that $m(\emptyset) > 0$ and e.g. for a classification problem of $k$ classes $m(\emptyset) > \frac{1}{k}$ (the evaluated model should be better than random guessing).

This definition can produce negative values, as well.
In fact, $v(S)$ is in range $[-1, 1]$ instead of $[0,1]$ \cite{cardin2007non}.
As stated in \cite{cardin2007non}, ``the meaning of the sign is clear.
For positive values, the corresponding criterion has to be considered, in average, as a benefit, conversely, for negative values, it represents a cost''.

As another example, Leon \cite{Leon14} uses a compound metric of the correlation coefficient and an error measure but does not explain this choice in-depth.

\ifthenelse{\boolean{reduced_version}}{
	% No mention of perspective on layers in reduced version
}{
	Figure \ref{fig:perspectives} visualizes two possible homogeneous perspectives for transforming an artificial neural network model into a coalitional game.
	The first perspective is exhaustively used in the experiment section: neurons within one layer of a multi-layer feed-forward neural network are used as players of the coalitional game.
	Temporarily removing single neurons from an inference step leads to the evaluation $v(S)$ of a subset $S$ of players $U$.

	Analogously (b) sketches a perspectives on layers as players of the game.
	To preserve the function of the overall network when single layers are omitted, skip-layer connections\footnote{Often also called residual connections as in \cite{he2016deep}} between layers are used.
	Selecting a subset $S$ of players then defines which layers to omit on the evaluation of $v(S)$.

	\begin{\figureType}[htb]
		\centering
		\subfloat[]{\includegraphics[height=3cm]{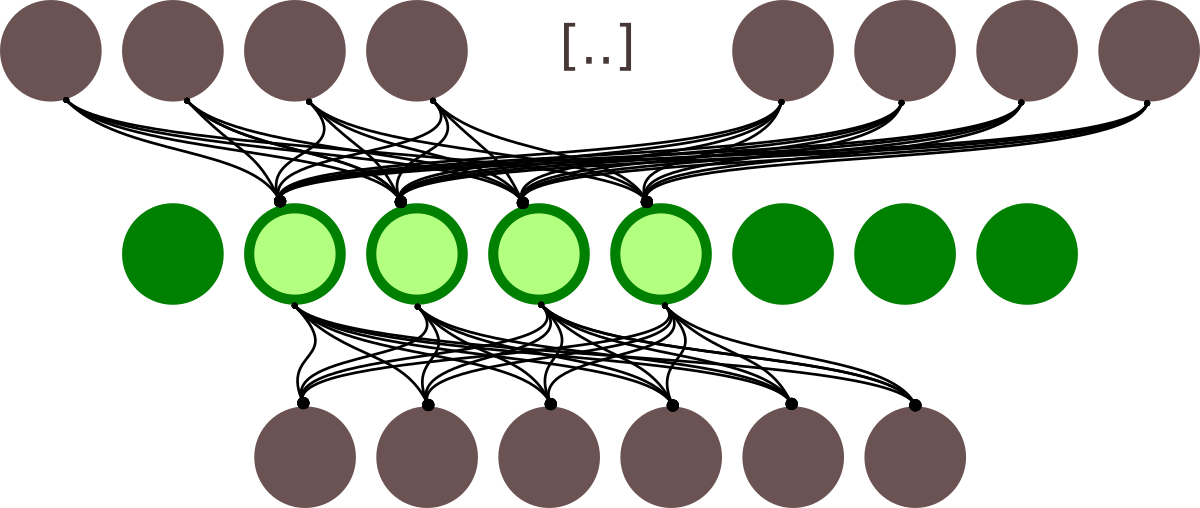}}
		\hfill
		\subfloat[]{\includegraphics[height=3cm]{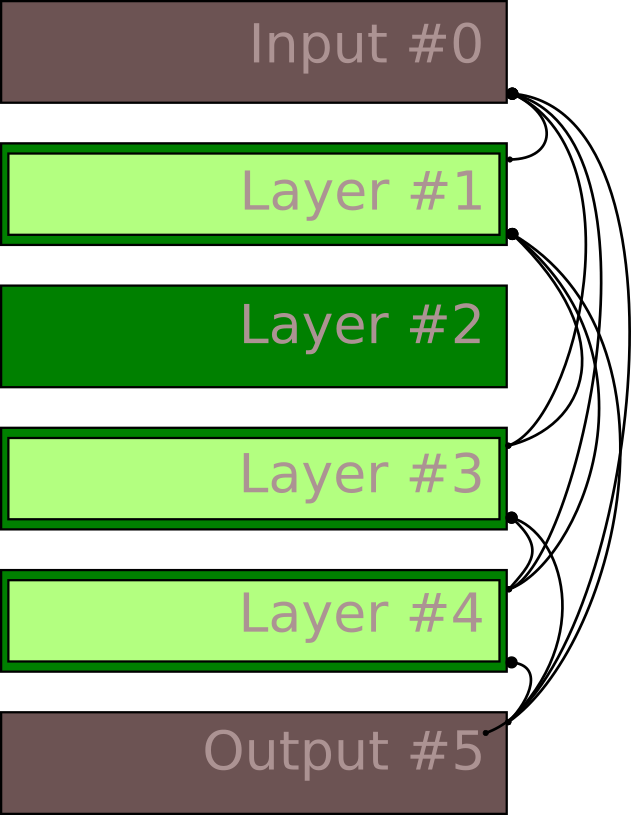}}
		\caption{
			\textit{(a)} A single hidden layer neural network with a game perspective on neurons: players $U$ are neurons in the second (or hidden) layer and a subset $S$ is highlighted to sketch the evaluation of $v(S)$.
			\textit{(b)} Perspective on layers as players $U$ of the game. Connections from and to layer \#2 are left out to evaluate $v(S)$.
		}
		\label{fig:perspectives}
	\end{\figureType}
}

%
% Strategies
%
\subsection{Strategies to obtain network topologies}
A top-down method derives a neural network structure from an initial (potentially large) root model in a \textit{derivation process}.
The result of such a process is a trained ANN model with potentially as few players (e.g. neurons) as possible, but at least fewer players than the initial root model.
One important step of the derivation process is the pruning of concrete structural components, which depends on the chosen \textit{pruning strategy}.
%The whole derivation process follows a certain strategy. % this is a non-content sentence. Clearly it follows a strategy, but which one?
%Hence, one can discriminate between different.
Each strategy defines a derivation rule, optional requirements\footnote{Requirements of a strategy include  the computation of statistical values such as weight norm for $Wbottom(k)$ or \termSV for $SVbottom(p)$.} and a stopping criterion.
The derivation rule selects a set of players to remove in each derivation step.
The stopping criterion decides if further pruning is possible or if the overall process terminates.

This work examined three different families of strategies: random-based, weight-based and Shapley-value-based strategies.
An overview of strategies is listed in table \ref{tbl:pruning-strategies}.

\begin{table*}[tb]
	\centering

	\begin{tabular}{ll}
		\hline
		\multicolumn{2}{c}{Pruning Strategies} \\
		\hline
		\hline
		Name & Description \\
		\hline
		$SVbottom(k)$ & Prune $k$ players with smallest \termSV.\\
		$SVbottom(p)$ & Prune players with \termSV below $p$: $\phi(i) < p\cdot\frac{1}{|S|}$.\\
		$SVbucket(p)$ & Prune $n$ players with smallest \termSVs such that $\sum\limits_{i=0}^n \phi(i) < p$.\\
		\hline
		$random(k)$ & Prune $k$ random players.\\
		\hline
		$Wbottom(k)$ & Prune $k$ players with smallest norm of their weights;\\
					& e.g. $\text{norm}(n) = \sqrt{\sum_{j\in E^{in}_n} (w_{j,n})^2}$.\\
		~
	\end{tabular}
	\caption[Pruning strategies analysed in this work]{
		Pruning strategies gathered and analysed in the course of this work.
		The strategy used by Leon \cite{Leon14} is $SVbottom(p)$, but with differences in technical detail.
	}
	\label{tbl:pruning-strategies}
\end{table*}

Strategies which select a fixed number of players for pruning within one step can be considered na{\"\i}ve or non-dynamic.
They mostly prune too few players in early steps and too many in late steps with few players left.
\\

Strategies based on random selections with a fixed size are an example for such na{\"\i}ve strategies.
A random-based strategy with $k$ players prunes $k$ randomly selected players.
Any strategy claiming to use a well-founded heuristic to select players for pruning must compete against random selection.
\\

A lot of existing strategies are based on information of the network's weights.
Three of such weight-based strategies are analysed in Wang et al \cite{wang2017towards}.
Their first strategy is based on ``$\sigma(R)$ score [..] generalized from'' approaches gathered by Thimm et al \cite{thimm1995evaluating}.
Those gathered approaches include smallest weight pruning \textit{min(w)} and sensitivity of a network to removal of a weight by monitoring sum of all weights changes during training.
The third strategy of Wang et al ``uses the average value of absolute weights sum'' of a neuron.

Here, a na{\"\i}ve pruning strategy based on weights is chosen as baseline comparison.
The strategy prunes $k$ players with the smallest norm of their weights.

For a given neuron $n$ its norm is calculated as $\text{norm}(n) = \sqrt{\sum_{j\in E^{in}_n} (w_{j,n})^2}$.
with $E^{in}_n$ being the set of incoming connections to neuron $n$ and $w_{j,n}$ the weight for the connection from neuron $j$ to neuron $n$.

If the game perspective defines one player as one neuron, simply norm$(n)$ can be considered for the strategy.
For multiple neurons treated as one player, one might sum this norm over all neurons of the concerned player.
With a fine-grained perspective on weights, one can use the related weights of the player for the root of summed, squared weights.
\\

Based on the game design given in section \ref{sec:design} three types of pruning strategies based on \termSVs are proposed:
A na{\"\i}ve strategy is given as $SVbottom(k)$ which prunes $k$ players with lowest \termSV in analogy to $random(k)$ and $Wbottom(k)$.

More dynamically, $SVbottom(p)$ prunes a player $i$ given its \termSV is below a threshold given by factor $p$ and the current average contribution if it would be uniformly distributed:
\[
	\phi(i) < p\cdot\frac{1}{|S|}
\]
This approach is similar to the one used by Leon in which ``the maximum Shapley value threshold to eliminate network elements is $\theta = \theta_s\cdot a_s$ where $\theta_s \in \{ 0, 0.1, 0.25 \}$ and $a_s$ is the average Shapley value of all existing network elements.''\cite{Leon14}.
The dynamic threshold in both methods can be considered highly similar as the expectation of \termSVs evidently matches the expectation of a uniform distribution.
Computing the expectation $\frac{1}{|S|}$ is neat and advantages of using an average of approximated \termSVs could not be found.
Like in the method of Leon, if no player below this threshold is found, ``the one with the minimum value becomes the candidate for elimination.''\cite{Leon14}.

The third strategy, $SVbucket(p)$, prunes players $i\in T$ such that $\underset{T\subset U}{\mathrm{argmax}}(|T|)$ and
\[
	\sum\limits_{i\in T} \phi(i) < p
\]
In other words, $SVbucket(p)$ collects all players $i \in T$ with smallest \termSV as long as their sum of \termSVs stays below a bucket value $p$.
Again, if no player matches this criterion, the one with the smallest \termSV is selected.

%
% Experiment: Shapley value approximation analysis
%
\subsection{Approximating the \termSV}
Calculating the exact \termSV requires averaging over all $2^N$ possible coalition, which is computational too expensive.
In fact, it is NP-complete \cite{deng1994complexity}.
To overcome this limitation, the \termSV is usually approximated through random sampling, as proposed in \cite{fatima2007randomized}.
Because there has been recent focus on other methods for approximating the \termSV such as sampling-based polynomial calculations \cite{castro2009polynomial} or structured random sampling \cite{hamers2016new} and to identify the applicability of random sampling for our approach, we condcuted preliminary experiments with randomly generated coalitional games (\textit{Randomly Pertubated Uniform game}) and the well-known \textit{United Nations Security Council game} \cite{roth1988}.
While the first games evaluate the approximation errors in case of almost uniformly distributed contributions with small perturbations, the second game addresses a non-uniform distribution of contributions.
Our experimental results show, that  at least 100 random samples are required for the permutation definition and at least 500 random samples are required for the subset definition.
The results go along with experiments in \cite{fatima2007randomized} in which in ``most cases, the error is less than 5\%.''.
Due to space constraints, we do not present further details here.

%
% Experiments
%
\section{Experiments}\label{sec:experiments}
% Overview of the section containing experiments
To assess the expressiveness of \termSVs we used them in context of pruning and compared them to methods with different heuristics.
For this assessment, we conducted the following experiments:
\begin{enumerate}
	\item \textbf{MNIST pruning} Pruning \textit{MNIST} models in an iterative top-down manner based on different strategies including ones based on \termSVs.
	\item \textbf{Pruning evaluation} An evaluation of \termSVbased pruning by comparing it with random selections of models obtained by grid search.
	\item \textbf{20newsgroups pruning} Pruning of larger \textit{20newsgroups} models for comparison with previous insights and proof of scale.
\end{enumerate}

In order to understand the effect of the \termSV, we stated several questions which can be summarized as following:
\begin{itemize}
	\item Which strategy requires the least number of steps?
	\item Can we define a lower bound for the game size given a problem and a threshold for the evaluation measure?
	\item How stable are the examined strategies?
\end{itemize}
The first question addresses \textit{how many steps} each strategy requires before the performance drops below a given threshold $\theta \in (0,1)$ for the evaluation measure.
The number of steps directly influences the overall number of training epochs and thus reflects a computational cost.
Given $\theta$, we also looked at which strategy found the \textit{least number of players} and if it could find this minimum repeatedly.
The found minimum across several strategies was compared to an exhaustive grid search (\textit{pruning evaluation}) to assess if there can be better performing models with less players found.
Strategies were calculated repeatedly to estimate their stability in terms of expectation and their standard deviation.
We did not only compare the found minimums, but also watched the pruning strategies along each step to compare how much contributional value (sum of \termSVs) was removed by non-Shapley-value-based strategies and if there can be any patterns found.

In the following, we briefly outline the experimental setup and results obtained for the experiments.

%
% Experiment: MNIST pruning
%
\subsection{\textit{MNIST} models pruning}
In the first experiment we evaluate our approach on a feed-forward network with a single hidden layer of size $h$ trained on the  MNIST \cite{lecun1998mnist} dataset using cross-entropy error.
Each neuron in the hidden layer is represented by a player in $U$ such that $|U| = h$.
We choose an initial hidden layer size $h = 40$ which forms our root-model.
After training the root-model for $T_0 = 20$ epochs, we apply the different pruning strategies outlined above.
Estimating the payoff $v(S)$ over a coallition of players, we calculate the accuracy over the test set for this coalition, thereby removing all neurons not in $S$. \\

Based on the selected pruning strategy we remove the least-contributing neurons resulting in a new model.
The new model is trained for $T_1 = 2$ epochs in order to compensate for the removed neurons.
We measure the validation set accuracy of the new model as estimator for the goodness of the pruning strategy.
If important neurons would have been pruned, we expect a lower accuracy than when pruning unimportant nodes.\\
Each experiment for a single pruning strategy is repeated independently twenty times in order to account for random effects.

\subsubsection{Results}
Figures \ifthenelse{\boolean{reduced_version}}{
	\ref{fig:pruning-walk-svbucket-vs-random} and \ref{fig:pruning-walk-svbucket-vs-weights} 
}{
	\ref{fig:pruning-walk-svbucket-vs-random}, \ref{fig:pruning-walk-svbucket-vs-weights} and \ref{fig:pruning-walk-svbucket-vs-svbottom}
} depict pruning walks of selected \termSVbased strategies in comparison with other strategies from \ref{tbl:pruning-strategies}.

The number of required steps for a single strategy is not directly viewable in this visualization.
For this, take a look at the listed tables such as table \ref{tbl:strategy-statistics-svbucket}.
Strategies with a fixed number of players to prune in each step, the number of steps can be directly obtained by dividing the number of initial players with the number of pruned players.
In case of a model-search with an accuracy threshold this number of steps does not apply as the strategy reaches the stopping criterion earlier.

Almost all strategies perform pretty similar for a number of players larger than twelve.
This fact is one indicator for the conclusion that contributions of a single player can be taken over by other players as long as the capacity of the network suffices to solve the problem.
A shift in the internal solving methodology of the network can not be identified.

Continuous lines depict the accuracy curve of each model.
Dotted lines are their respective sum of removed \termSV.
\\

Figure \ref{fig:pruning-walk-svbucket-vs-random} compares \textbf{SVbucket($0.2$)} with \textbf{random($1$)} and \textbf{random($3$)}.
The average number of steps for SVbucket($0.2$) is 12.05 with a minimum of 12 and a maximum of 13 steps.
Despite the algorithmic approach of strategies, randomness of the derivation process of obtaining and retraining a smaller model and the approximation error for \termSVs reflect slightly in these values.
Random($k$) have 40 and 14 fixed steps, respectively.

It can be observed that below twelve players the \termSVbased strategy is able to stay above random pruning.
Note, how the removed \termSV of each strategy stays below a threshold of 0.2 or increased with decreasing number of players, respectively.
\\

\ifthenelse{\boolean{condensed_image_version}}{
	% Condensed version with two figures in a row with two subfloats 2a & 2b
	\begin{figure*}[h!]
		\centering
		\subfloat[
			\textbf{SVbucket($0.2$) vs. random($k$)}:
				fewer steps and superior in moment of sudden accuracy decay:
				\termSVbased pruning is clearly preferrable to random guessing.
			\label{fig:pruning-walk-svbucket-vs-random}
		]{\includegraphics[width=0.45\textwidth]{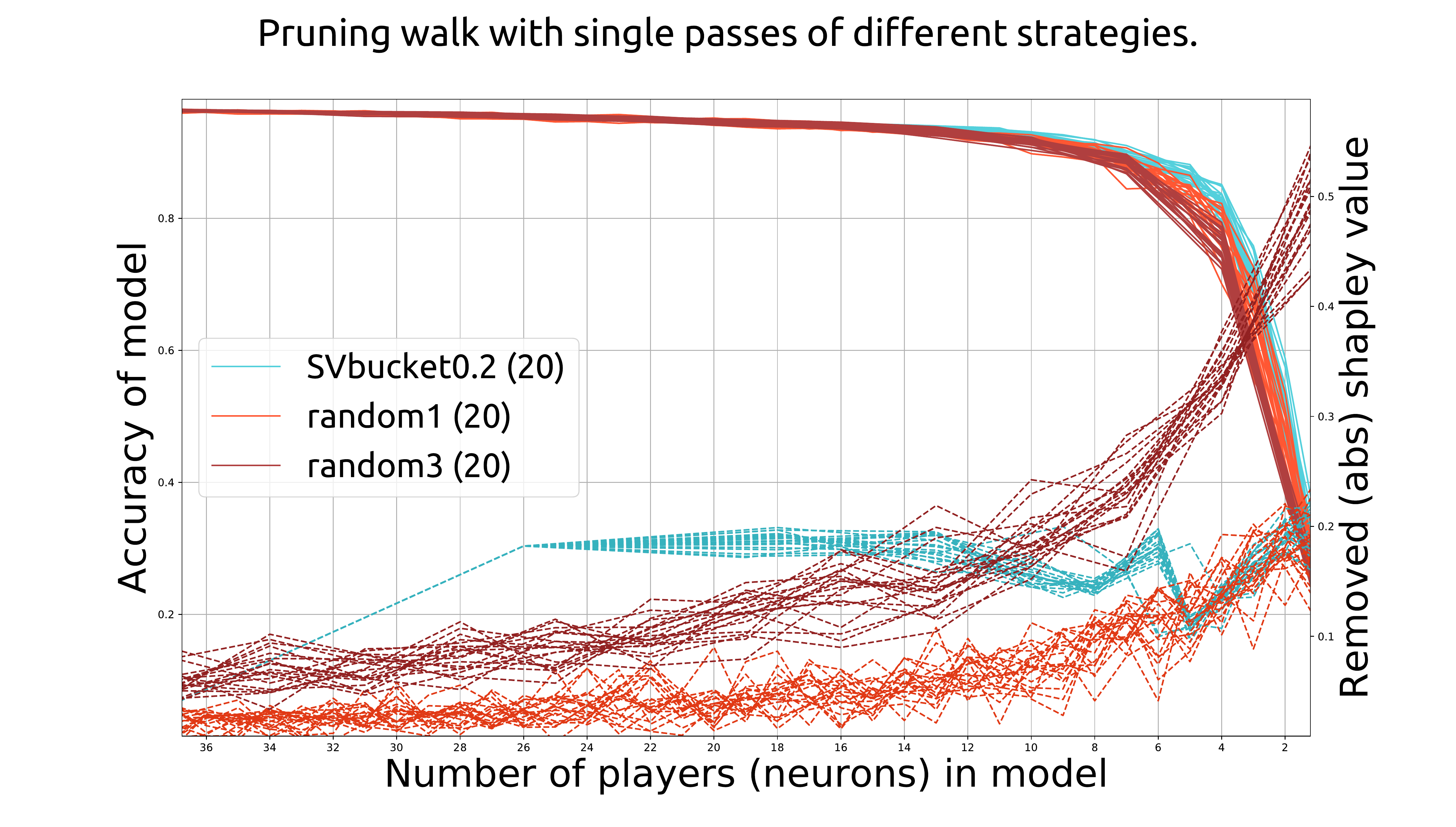}}
		~
		\subfloat[
			\textbf{SVbucket($0.2$) vs Wbottom($k$):}
				Weights are obviously no direct indicator for well contributing players.
				At least when considering pruning, a strategy based on weights is inferior to one based on \termSVs.
			\label{fig:pruning-walk-svbucket-vs-weights}
		]{\includegraphics[width=0.45\textwidth]{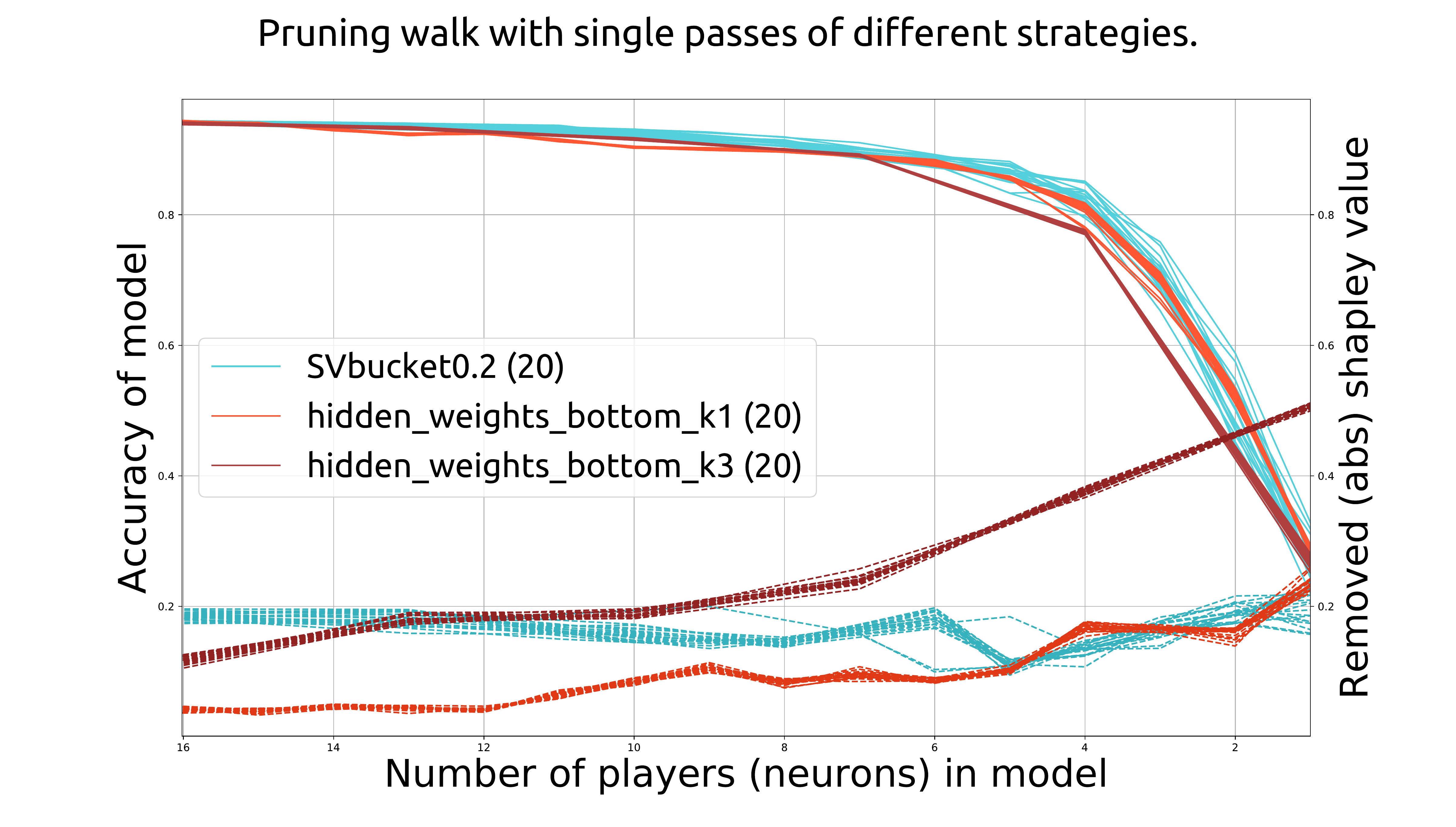}}
		\caption{Exemplary pruning walks. Comparing $SVBucket(p)$ with $random(k)$ and $Wbottom(k)$.}
	\end{figure*}
}{
	% Non-condensed single-image 2a
	\begin{figure*}[h!]
		\centering
		\includegraphics[width=\textwidth]{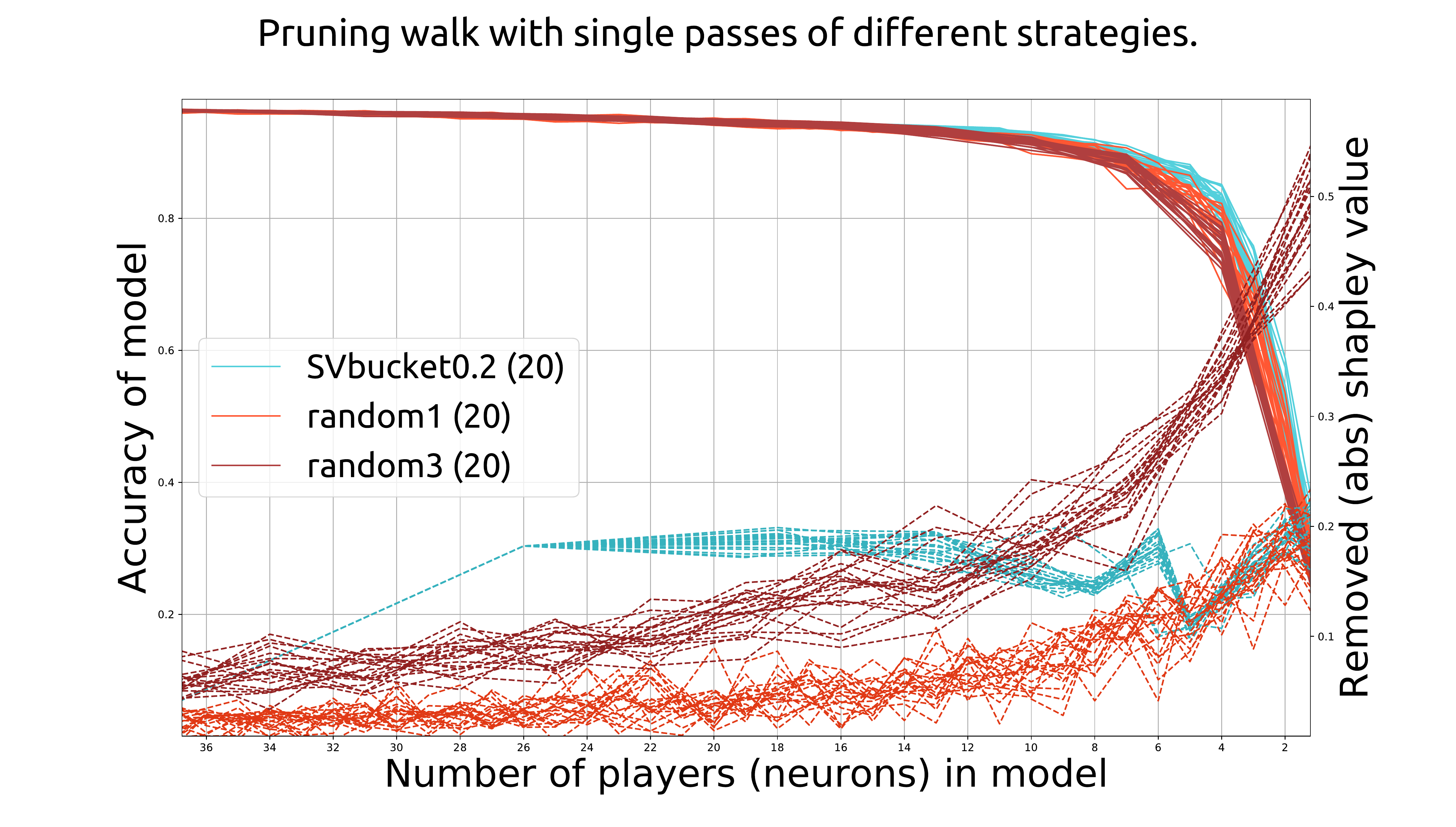}
		\caption{
			\textbf{SVbucket($0.2$) vs. random($k$)}:
			fewer steps and superior in moment of sudden accuracy decay:
			\termSVbased pruning is clearly preferrable to random guessing.
		}
		\label{fig:pruning-walk-svbucket-vs-random}
	\end{figure*}
}

Figure \ref{fig:pruning-walk-svbucket-vs-weights} compares \textbf{SVbucket($0.2$)} with \textbf{Wbottom($1$)} and \textbf{Wbottom($3$)}.
The weight-based strategies take 40 and 14 fixed steps, respectively.
It can be clearly seen that the \termSVbased strategy stays above the weighted-based one.
A weight-based strategy is questionable as low weights are assumably no indicator for contributional value of a single player.
\\

\ifthenelse{\boolean{condensed_image_version}}{
	% Condensed double-image 2a&2b is above
}{
	% Non-condensed single-image 2b
	\begin{figure*}[htb]
		\centering
		\includegraphics[width=\textwidth]{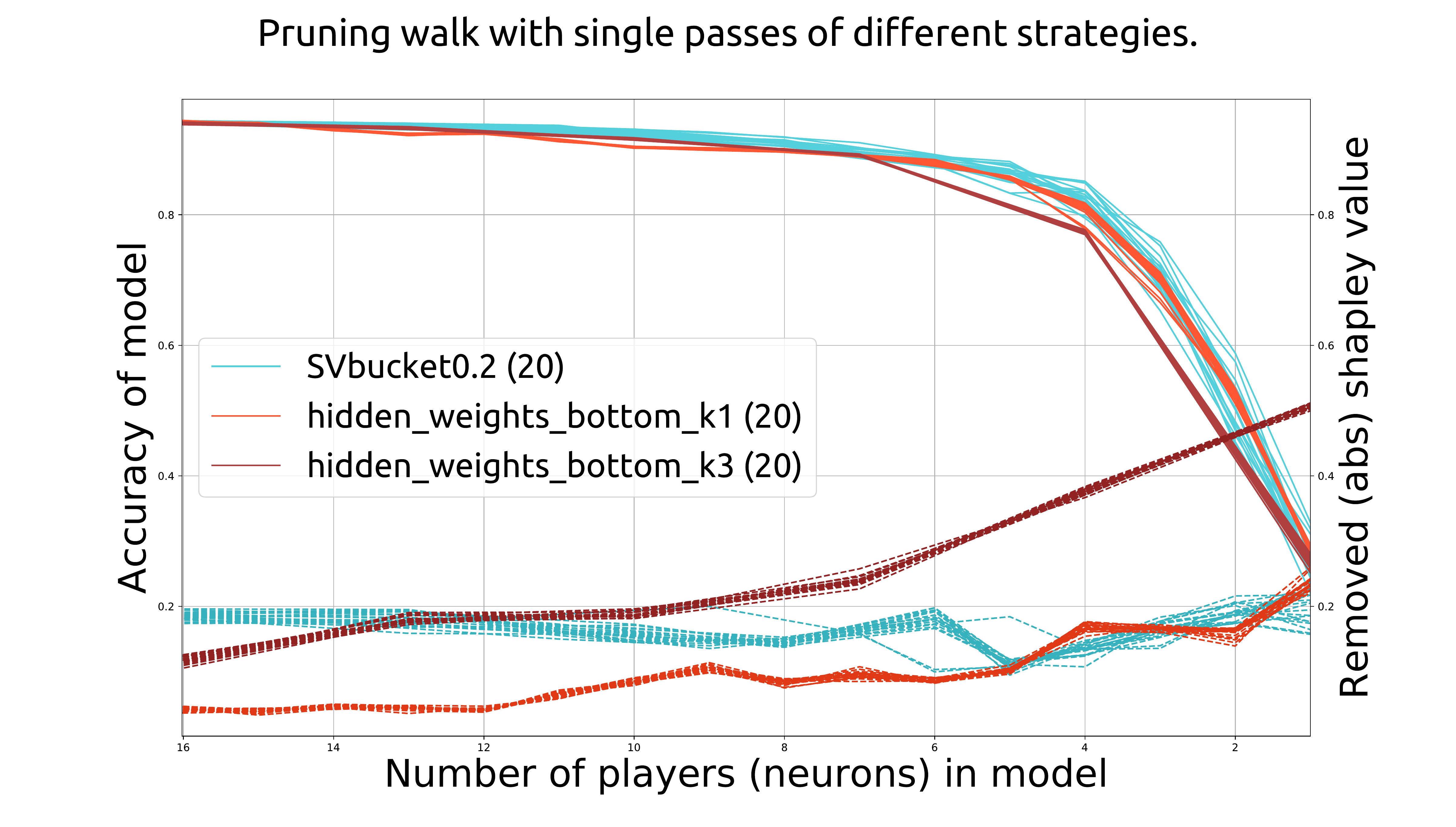}
		\caption{
			\textbf{SVbucket($0.2$) vs Wbottom($k$):}
			Weights are obviously no direct indicator for well contributing players.
			At least when considering pruning, a strategy based on weights is inferior to one based on \termSVs.
		}
		\label{fig:pruning-walk-svbucket-vs-weights}
	\end{figure*}
}

\ifthenelse{\boolean{reduced_version}}{
	% No pruning-walk-comparison between svbucket and svbottom in reduced version
}{
	Figure \ref{fig:pruning-walk-svbucket-vs-svbottom} compares \textbf{SVbucket($0.2$)} with \textbf{SVbottomP($0.5$)}.
}
SVbottomP($0.5$) takes 31.35 steps on average, its maximum is 33 and its minimum 30.
While SVbucket($0.2$) starts with pruning a larger amount of players (fitting into a bucket of $p = 0.2$) SVbottomP($0.5$) slightly increased the amount of \termSV to be removed in each step as the average \termSV increases.
Before the moment of sudden accuracy decay (below eight players) one could argue SVbottomP($0.5$) to be superior to SVbucket($0.2$) -- it prunes fewer players.
However, it also takes more steps and SVbucket($0.2$) is able to jump pretty far down within a few steps which saves a lot of retraining epochs.
\\

\ifthenelse{\boolean{reduced_version}}{
	% No pruning-walk-comparison between svbucket and svbottom in reduced version
}{
	\ifthenelse{\boolean{condensed_image_version}}{
		% Condensed single-image
		\begin{figure}[h!]
			\centering
			\includegraphics[width=0.45\textwidth]{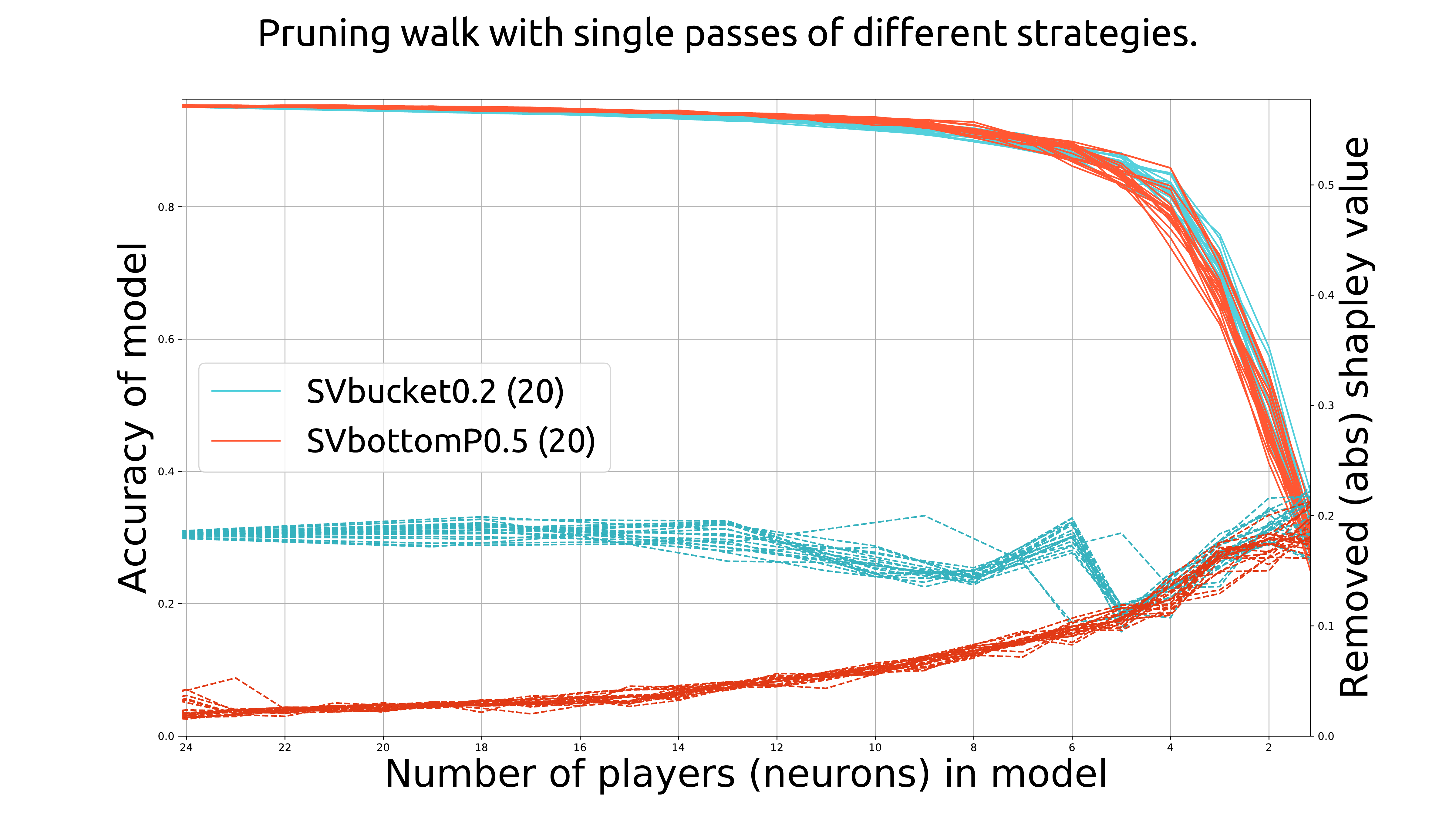}
			\caption{
				\textbf{SVbucket($0.2$) vs SVbottomP($0.5$):}
				On average, SVbottomP($0.5$) requires more than double the times (31.35) of training epochs than SVbucket($0.2$) (12.05).
				The accuracy curve for SVbottomP($0.5$) with models below six players shows higher variance in comparison.
				It can be argued to prefer a cumulating strategy such as SVbucket($0.2$) over working with the average \termSV as in SVbottomP($0.5$) or as proposed by Leon \cite{Leon14}.
			}
			\label{fig:pruning-walk-svbucket-vs-svbottom}
		\end{figure}
	}{
		% Non-condensed single-image
		\begin{figure*}[h!]
			\centering
			\includegraphics[width=\textwidth]{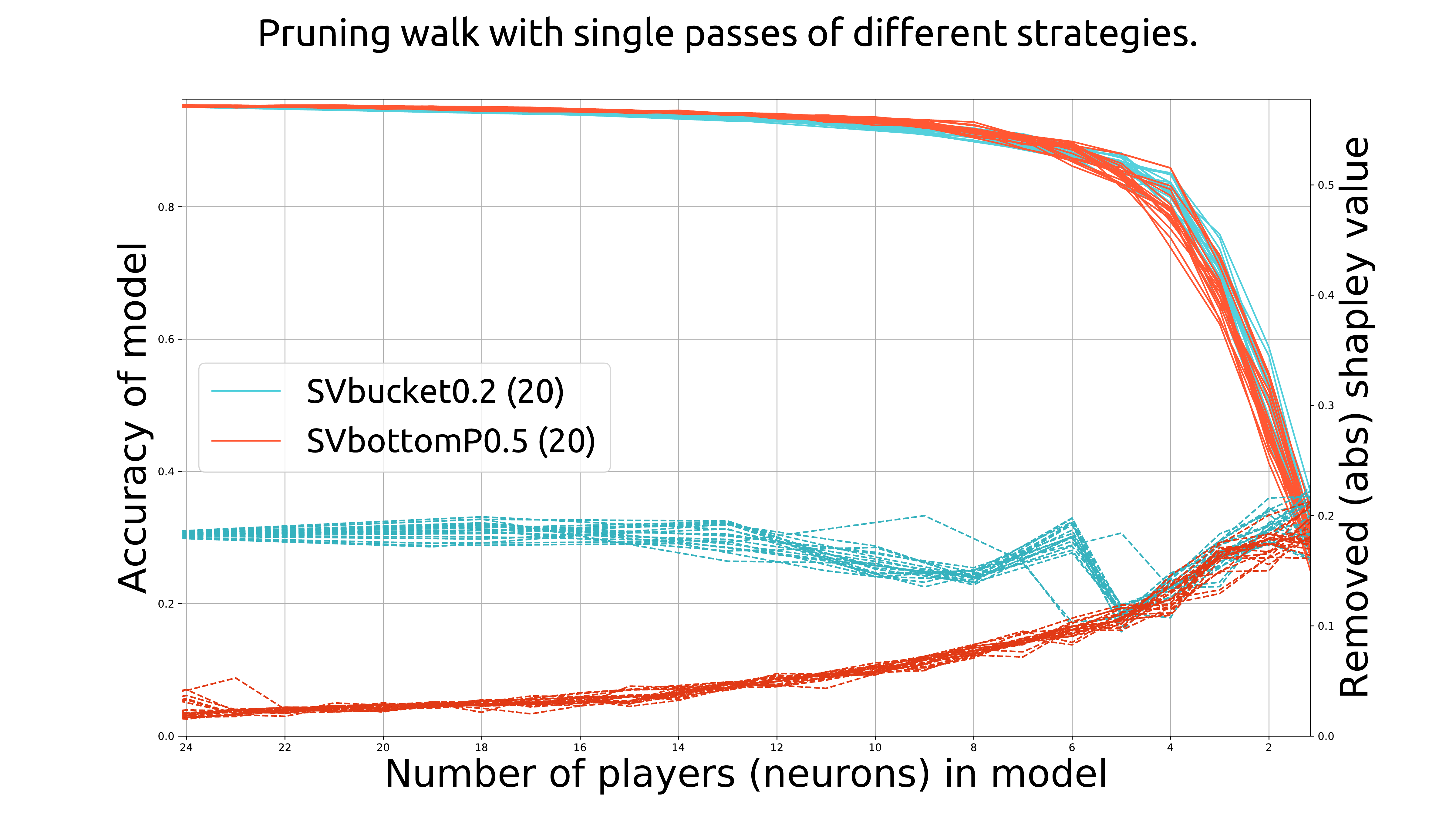}
			\caption{
				\textbf{SVbucket($0.2$) vs SVbottomP($0.5$):}
				On average, SVbottomP($0.5$) requires more than double the times (31.35) of training epochs than SVbucket($0.2$) (12.05).
				The accuracy curve for SVbottomP($0.5$) with models below six players shows higher variance in comparison.
				It can be argued to prefer a cumulating strategy such as SVbucket($0.2$) over working with the average \termSV as in SVbottomP($0.5$) or as proposed by Leon \cite{Leon14}.
			}
			\label{fig:pruning-walk-svbucket-vs-svbottom}
		\end{figure*}
	}
}

In search for a minimum number of required players to stay above an accuracy threshold of $\theta = 0.9$ some statistical values for SVbucket($p$) and SVbottomP($p$) are given in tables \ref{tbl:strategy-statistics-svbucket} and \ref{tbl:strategy-statistics-svbottomp}.
Average number of steps (obtained model versions) and possible outliers in those values are easy to compare runtimes required for a grid search conducted in the following experiment.
It can be clearly seen that one of the described top-down strategies is able to find a value near the minimum within a significant lower required training epochs.

\ifthenelse{\boolean{wide_tables}}{
	%
	% Two tables combined to one wide table
	%
	\begin{\reducibleTableType}[tb]
		\centering

		\subfloat[A low bucket value of 0.1 for pruning requires a lot of steps.]{
\begin{tabular}{ll}
	\multicolumn{2}{c}{SVbucket0.1} \\
	\hline
	avg \# steps & 19.65 \\
	max \# steps & 21 \\
	min \# steps & 19 \\
	avg \# epochs & 59.3 \\
	max \# epochs & 62 \\
	min \# epochs & 58 \\
	avg \# found & 6.95 \\
	max \# found & 8 \\
	min \# found & 6 \\
	\hline
\end{tabular}
			\label{tbl:strategy-statistics-svbucket}
		}
		~
		\subfloat[A compromise between pruning too fast and still having few retraining epochs.]{
\begin{tabular}{ll}
	\multicolumn{2}{c}{SVbucket0.2} \\
	\hline
	avg \# steps & 12.05 \\
	max \# steps & 13 \\
	min \# steps & 12 \\
	avg \# epochs & 44.1 \\
	max \# epochs & 46 \\
	min \# epochs & 44 \\
	avg \# found & 8.05 \\
	max \# found & 10 \\
	min \# found & 7 \\
	\hline
\end{tabular}
		}
	\ifthenelse{\boolean{reduced_version}}{
		% No statistics for SVbucket(0.3) in reduced version
	}{
		~
		\subfloat[A large bucket of 0.3 for pruning allows to reduce the steps.]{
\begin{tabular}{ll}
	\multicolumn{2}{c}{SVbucket0.3} \\
	\hline
	avg \# steps & 8.0 \\
	max \# steps & 8 \\
	min \# steps & 8 \\
	avg \# epochs & 36.0 \\
	max \# epochs & 36 \\
	min \# epochs & 36 \\
	avg \# found & 9.4 \\
	max \# found & 12 \\
	min \# found & 8 \\
	\hline
\end{tabular}
		}
	}
		~
		\subfloat[SVbottomP0.5]{
\begin{tabular}{ll}
	\multicolumn{2}{c}{SVbottomP0.5} \\
	\hline
	avg \# steps & 31.35 \\
	max \# steps & 33 \\
	min \# steps & 30 \\
	avg \# epochs & 82.7 \\
	max \# epochs & 86 \\
	min \# epochs & 80 \\
	avg \# found & 7.25 \\
	max \# found & 8 \\
	min \# found & 7 \\
	\hline
\end{tabular}
			\label{tbl:strategy-statistics-svbottomp}
		}
		~
		\subfloat[SVbottomP0.7]{
\begin{tabular}{ll}
	\multicolumn{2}{c}{SVbottomP0.7} \\
	\hline
	avg \# steps & 15.75 \\
	max \# steps & 19 \\
	min \# steps & 13 \\
	avg \# epochs & 51.5 \\
	max \# epochs & 58 \\
	min \# epochs & 46 \\
	avg \# found & 7.25 \\
	max \# found & 8 \\
	min \# found & 6 \\
	\hline
\end{tabular}
		}
		\caption[Statistics for SVbucket\&SVbottom-strategy]{
			\textbf{SVbucket}-strategy:
			The table shows average, maximum and minimum number of steps to reach the stopping criterion (no more neurons to prune).
			It also shows statistics for the total number of required epochs during the destructive iterative approach.
			The average, maximum and minimum number of neurons to stay above the threshold obtained by all repeated strategies is denoted as ``found''.

			Starting with a pre-trained root model with 20 training epochs, SVbucket results in an almost constant number of required epochs.
			Depending on the used parameter, the strategy finds an average of seven, eight or nine minimum number of players to stay above a threshold of 0.9 in accuracy within a total maximum number of required training epochs of 60, 45 and 36.

			\textbf{SVbottomP}-strategy: With a total number of required training epochs of 52 and 83 the strategy is able to find a minimum number of seven required players to stay above a threshold of 0.9 in accuracy for classifying MNIST.
			Twenty repeated runs are performed for each strategy to obtain the statistics and confirms the stability of the method.
		}
	\end{\reducibleTableType}
}{
	%
	% Two separated thin tables
	%
	\begin{\reducibleTableType}[tb]
		\centering

		\subfloat[A low bucket value of 0.1 for pruning requires a lot of steps.]{
			\begin{tabular}{ll}
	\multicolumn{2}{c}{SVbucket0.1} \\
	\hline
	avg \# steps & 19.65 \\
	max \# steps & 21 \\
	min \# steps & 19 \\
	avg \# epochs & 59.3 \\
	max \# epochs & 62 \\
	min \# epochs & 58 \\
	avg \# found & 6.95 \\
	max \# found & 8 \\
	min \# found & 6 \\
	\hline
\end{tabular}

		}
		~
		\subfloat[A compromise between pruning too fast and still having few retraining epochs.]{
			\begin{tabular}{ll}
	\multicolumn{2}{c}{SVbucket0.2} \\
	\hline
	avg \# steps & 12.05 \\
	max \# steps & 13 \\
	min \# steps & 12 \\
	avg \# epochs & 44.1 \\
	max \# epochs & 46 \\
	min \# epochs & 44 \\
	avg \# found & 8.05 \\
	max \# found & 10 \\
	min \# found & 7 \\
	\hline
\end{tabular}

		}
	\ifthenelse{\boolean{reduced_version}}{
		% No statistics for SVbucket(0.3) in reduced version
	}{
		~
		\subfloat[A large bucket of 0.3 for pruning allows to reduce the steps.]{
			\begin{tabular}{ll}
	\multicolumn{2}{c}{SVbucket0.3} \\
	\hline
	avg \# steps & 8.0 \\
	max \# steps & 8 \\
	min \# steps & 8 \\
	avg \# epochs & 36.0 \\
	max \# epochs & 36 \\
	min \# epochs & 36 \\
	avg \# found & 9.4 \\
	max \# found & 12 \\
	min \# found & 8 \\
	\hline
\end{tabular}

		}
	}

		\caption[Statistics for SVbucket-strategy]{
			\textbf{SVbucket}-strategy:
			The table shows average, maximum and minimum number of steps to reach the stopping criterion (no more neurons to prune).
			It also shows statistics for the total number of required epochs during the destructive iterative approach.
			The average, maximum and minimum number of neurons to stay above the threshold obtained by all repeated strategies is denoted as ``found''.

			Starting with a pre-trained root model with 20 training epochs, SVbucket results in an almost constant number of required epochs.
			Depending on the used parameter, the strategy finds an average of seven, eight or nine minimum number of players to stay above a threshold of 0.9 in accuracy within a total maximum number of required training epochs of 60, 45 and 36.
		}
		\label{tbl:strategy-statistics-svbucket}
	\end{\reducibleTableType}

	\begin{\tableType}[tb]
		\centering

		\subfloat[SVbottomP0.5]{
			\begin{tabular}{ll}
	\multicolumn{2}{c}{SVbottomP0.5} \\
	\hline
	avg \# steps & 31.35 \\
	max \# steps & 33 \\
	min \# steps & 30 \\
	avg \# epochs & 82.7 \\
	max \# epochs & 86 \\
	min \# epochs & 80 \\
	avg \# found & 7.25 \\
	max \# found & 8 \\
	min \# found & 7 \\
	\hline
\end{tabular}

		}
		~
		\subfloat[SVbottomP0.7]{
			\begin{tabular}{ll}
	\multicolumn{2}{c}{SVbottomP0.5} \\
	\hline
	avg \# steps & 31.35 \\
	max \# steps & 33 \\
	min \# steps & 30 \\
	avg \# epochs & 82.7 \\
	max \# epochs & 86 \\
	min \# epochs & 80 \\
	avg \# found & 7.25 \\
	max \# found & 8 \\
	min \# found & 7 \\
	\hline
\end{tabular}

		}
		\caption[Statistics for SVbottom-strategy]{
			\textbf{SVbottomP}-strategy: With a total number of required training epochs of 52 and 83 the strategy is able to find a minimum number of seven required players to stay above a threshold of 0.9 in accuracy for classifying MNIST.
			Twenty repeated runs are performed for each strategy to obtain the statistics and confirms the stability of the method.
		}
		\label{tbl:strategy-statistics-svbottomp}
	\end{\tableType}
}

%
% Experiment: Grid search (bottom up)
%
\subsection{\textit{MNIST} models grid search}
For hyperparameters \textit{number of training epochs} and \textit{number of players} a grid search is conducted and repeated 200 times.
Each result is an accuracy value obtained from a MNIST feedforward network, trained with the according number of epochs and number of given players (hidden neurons).

\ifthenelse{\boolean{reduced_version}}{
	% no grid-search table in reduced version
}{
	A visualization of 200 repetitions with each model being trained with 50 epochs is given in figure \ref{fig:grid-search-50-20-epochs}.
	Data with less than 20 training epochs shows significant lower accuracies on average.
	If a minimum number of players with at least $\theta=0.9$ in accuracy is searched at least two models have to be trained and compared.
	An exhaustive search actually takes much more models for comparison and demonstrating the fact that a certain number of players actually must be close to the required minimum.
	The required number of epochs sums up with the required lookups.
	A lookup is sketched in figure \ref{fig:grid-search-50-20-epochs} as the intersection of the threshold lines at $\theta=0.9$ for accuracy and a vertical line at player amount of seven.

	\begin{figure*}[htb]
		\centering
		\ifthenelse{\boolean{condensed_image_version}}{
			% Condensed single-image
			\includegraphics[width=0.8\textwidth]{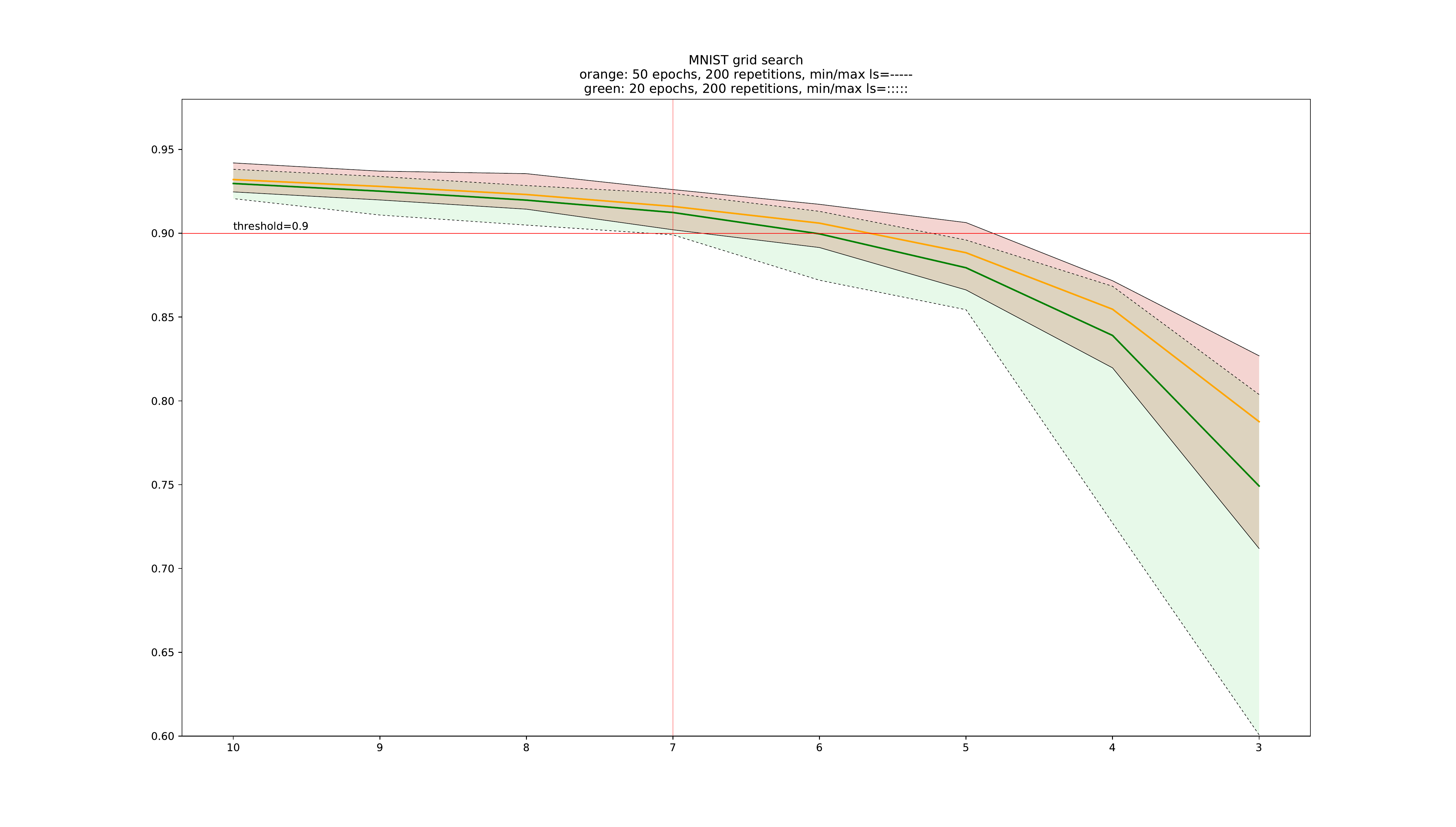}
		}{
			% Non-condensed single-image
			\includegraphics[width=\textwidth]{2017-10-17-mnist-grid-search-epochs50-20-reps200-thresholds.pdf}
		}
		\caption{
			\textbf{Grid search:}
			For $p\in[3,10]$ and $epochs\in[10,50]$ a grid search for MNIST-FFN was performed and accuracy values for the test set obtained.
			The number of training epochs reduce the accuracy variance of trained models between ten and 50 epochs.
			Grid search results suggest to use at least 20 or 30 number of training epochs before comparing different models and search for a minimum value for the number of players.
			Searching for a minimum number of players with an accuracy above an accuracy threshold $\theta$ sums up all epochs for each grid lookup.
			It is obvious that this is computationally more expensive than using one of the presented top-down pruning strategies.
		}
		\label{fig:grid-search-50-20-epochs}
	\end{figure*}
}

\ifthenelse{\boolean{reduced_version}}{
	% no grid-search table in reduced version
}{
	Some statistical results of the grid search are depicted in table \ref{tbl:grid-search}.
}

Comparing the grid search results to the MNIST experiments with different strategies, it gets obvious to state that a top-down pruning strategy is more efficient under assumption of obtaining \termSVs in constant or small linear time.
In fact, approximating \termSVs with 500 samples only takes 500 inference steps which are computationally cheap in comparison to training epochs.
Unlike training, approximating \termSV can be fully distributed and the parallel inference computations only need to be summed.
This parallelization is used in the underlying framework.
Grid search finds comparable models and minimum number of players with a technically more simple method.
With means of computational costs however, contribution-based pruning methods are faster and might provide more insights into ANNs in future.

%
% Experiment: 20newsgroups
%
\subsection{\textit{20newsgroups} models pruning}
The third experiment is set up equivalent to the first but based on the \textit{20newsgroups} dataset \cite{lang1995newsweeder}.
Evaluation measures are not directly comparable to MNIST.
However, statistical results for all strategies of the MNIST experiment could be reproduced.
The accuracy threshold was set to $\theta = 0.72$.
Repetitions of SVbucket($0.2$) produced an average of 9.2 players required in minimum for $\theta$.

%
% Future work
%
\section{Future Work}\label{sec:future-work}
\ifthenelse{\boolean{reduced_version}}{
	% Reduced version of future work
	\textbf{Theoretical aspects:}
	1) Properties like symmetry or efficiency might contribute to \textit{reducing the complexity} of computing Shapley values.
	2) The classic \termSV implies that all players of a game can form a meaningful coalition.
	Concerning neural networks, you might not have sets where every subset is meaningful.
	The game could be extended to account for network and not just coalition structure.
	Myerson \cite{myerson1977graphs} augmented the classic cooperative game by adding a network structure and obtained a \textit{communication game} in which the role of the network defines which coalitions can actually operate.

	\textbf{Algorithmic aspects:}
	1) Destructive, constructive or hybrid approaches could be navigated by \termSVs.
	2) Simulated annealing suggests that greedy steps can get you stuck in local minima.
	Mixing random pruning and \termSV guided pruning would be an example of another heuristic for topology optimization.
}{
	% More details on future work
	On the theory of coalitional games in artificial neural networks:
	\begin{itemize}
		\item Properties like symmetry or efficiency might contribute to reducing the complexity of computing Shapley values.
			Constructing a super-additive measure based on neural networks as payoff function for a coalitional game would make Shapley-value-based applications on neural networks more practical and could even lead to new applications like playing multiple games in parallel or combining existing ones.
		\item Comparing different evaluation measures as payoff of the game.
		\item The classic \termSV implies that all players of a game can form a meaningful coalition.
			Concerning neural networks, you might not have sets where every subset is meaningful.
			The game could be extended to account for network and not just coalition structure.
			Myerson \cite{myerson1977graphs} augmented the classic cooperative game by adding a network structure and obtained a \textit{communication game} in which the role of the network defines which coalitions can actually operate.
	\end{itemize}

	On algorithms based on Shapley value:
	\begin{itemize}
		\item Further destructive, constructive or hybrid approaches could be encouraged.
			Hybrid approaches could start with dozens of layers, switch to neurons and finally to weights -- based on different characteristics of the resulting network.
			This could lead to deep and sparse networks constructed top-down.
		\item Constructing networks bottom-up could also be navigated by Shapley values, e.g. strong contributing neurons could be duplicated or new network components could be added until some characteristic is achieved, e.g. a desired distribution of Shapley values.
		\item Simulated annealing suggests that greedy steps can get you stuck in local minima.
			A mix of random pruning and Shapley value guided pruning could define a new heuristics for topology optimization.
	\end{itemize}
}

%
% Conclusion
%
\section{Conclusion}\label{sec:conclusion}
Shapley value as a solution concept for coalitional games can be applied to ANNs to obtain values of contribution for components such as neurons.

The idea was separated into a game on topologies to obtain \termSVs as measuring values and pruning strategies as one possible application.
The methodology was generalised to different perspectives on neural networks.

Choices in course of constructing a game on topologies -- such as the perspective or the payoff function -- were discussed.
It was discovered that the properties symmetry, efficiency and additivity of the Shapley value can not easily be derived.
This led to further research questions, e.g. ``is it possible to construct a super-additive payoff function based on neural networks?''.

The question of usefulness of the Shapley value was approached within the scope of pruning methods for neural networks.
On the one hand, it could be shown that methods based on the Shapley value have a longer power of endurance in terms of their accuracy when pruned.
On the other hand, only models with already severely reduced components showed significant differences between strategies.

The hypothesis for this observation is that the analysed problems are still low in their complexity while very large models just offer a large amount of redundant degrees of freedom.
Reducing these parameters then, it can be observed that the model compensates the loss without dropping significantly in accuracy as long as there are still a large amount of parameters.
This goes along with findings in compressing neural networks via pruning, e.g. as in Han et. al \cite{han2015learning}.
Neural networks are able to compensate neural loss if they undergo a retraining phase.

\TermSVs give a hint of the importance of different network components.
However, the competence of such an component can be taken over by equivalent components when pruned.

\bibliographystyle{elsarticle-harv}
\bibliography{bibliography}

\ifthenelse{\boolean{reduced_version}}{
	% no grid-search table in reduced version
}{
\begin{table*}[tb]
	\centering

	\begin{tabular}{lllllllll}
		\hline
		\multicolumn{9}{c}{\textbf{MNIST grid search}} \\
		\#players & \multicolumn{1}{c}{3} & \multicolumn{1}{c}{4} & \multicolumn{1}{c}{5} & \multicolumn{1}{c}{6} & \multicolumn{1}{c}{7} & \multicolumn{1}{c}{8} & \multicolumn{1}{c}{9} & \multicolumn{1}{c}{10} \\
		\hline
		\hline
		\multicolumn{9}{c}{Used \textbf{15 epochs}, 100 repetitions} \\
		\hline
		Min & 0.6210 & 0.7258 & 0.8265 & 0.8310 & 0.8992 & 0.9040 & 0.9037 & 0.9207 \\
		Mean & 0.7243 & 0.8268 & 0.8750 & 0.8976 & 0.9101 & 0.9175 & 0.9236 & 0.9282 \\
		Max & 0.7904 & 0.8585 & 0.9010 & 0.9110 & 0.9177 & 0.9267 & 0.9315 & 0.9398 \\
		\hline
		\multicolumn{9}{c}{Used \textbf{20 epochs}, 100 repetitions} \\
		\hline
		Min & 0.6249 & 0.7271 & 0.8547 & 0.8838 & 0.8991 & 0.9058 & 0.9109 & 0.9229 \\
		Mean & 0.7525 & 0.8382 & 0.8792 & 0.8990 & 0.9120 & 0.9200 & 0.9251 & 0.9300 \\
		Max & 0.8038 & 0.8642 & 0.8960 & 0.9131 & 0.9212 & 0.9285 & 0.9338 & 0.9367 \\
		\hline
		\multicolumn{9}{c}{Used \textbf{30 epochs}, 100 repetitions} \\
		\hline
		Min & 0.6255 & 0.8108 & 0.8686 & 0.8831 & 0.9052 & 0.9111 & 0.9197 & 0.9251 \\
		Mean & 0.7677 & 0.8502 & 0.8853 & 0.9028 & 0.9148 & 0.9217 & 0.9278 & 0.9315 \\
		Max & 0.8176 & 0.8676 & 0.9052 & 0.9137 & 0.9241 & 0.9309 & 0.9367 & 0.9391 \\
		\hline
		\multicolumn{9}{c}{Used \textbf{50 epochs}, 100 repetitions} \\
		Min & 0.7288 & 0.8197 & 0.8662 & 0.8917 & 0.9076 & 0.9144 & 0.9199 & 0.9247 \\
		Mean & 0.7860 & 0.8543 & 0.8884 & 0.9062 & 0.9165 & 0.9229 & 0.9274 & 0.9323 \\
		Max & 0.8167 & 0.8717 & 0.9064 & 0.9173 & 0.9257 & 0.9356 & 0.9358 & 0.9403 \\
		\hline
	\end{tabular}
	\caption[MNIST grid search]{
		A summarizing excerpt of the exhaustive grid search with several hundred repetitions.
	}
	\label{tbl:grid-search}
\end{table*}
}

\ifthenelse{\boolean{with_rebuttal}}{\appendix
\section{Rebuttal}
Thank you for your review and the possibility to submit a last page containing a rebuttal.\\

\noindent {\color{gray}``The Authors do not provide explicit evidence on the contribution and novelty of their work, compared to previous similar works on the subject. Several other papers exist, which make use of the Shapley value for Artificial Neural Network topology optimization. The Authors are asked to provide a very solid support of the novelty and contribution of their work.''}\\
At first glance, the article could create doubt on novelty of the work considering the existing list of literature on \textit{Neural Networks} and \textit{Shapley Value}.
Concerning Artificial Neural Networks there has been a lot of work on destructive optimization (e.g. for better generalization, compression or performance).
Only one, namely the work of Leon \cite{Leon14}, introduced Shapley Value for pruning \textbf{Artificial} Neural Networks.
However, there exists previous work, as stated above, which already applied Shapley Values on \textbf{Biological} Neural Networks (mostly in research fields such as neurobiology, neuro-science, medicine).
Pruning with means of a measure of saliency is known since the 1990s but as of our knowledge there exists neither exhaustive comparative work of saliency measures nor much analytical insights about them.
Applying Shapley value as a measure of saliency for pruning Artificial Neural Networks can be considered as novel.

Compared to the work of Leon, this work separated Shapley Value conceptually from the idea of pruning Artificial Neural Networks and mentioned different possible perspectives in which contributional values for network components can be obtained.
Furthermore, the pruning methodology was extended to multiple strategies, which have been discussed and compared.
While Leon applied one concrete strategy to several simple problem domains, we applied different strategies to two larger domains, namely \textit{MNIST} and \textit{newsgroups20} (image and text classification), and discussed differences of those strategies.

We agree, that from a perspective of pruning Neural Networks, we did not contribute a novel method as it can be seen as pruning with a measure of saliency.
However, we showed a phase of long endurance in generalization for models pruned with Shapley Value.
The pruning justified using Shapley Value as a measure to inspect Artificial Neural Networks.

~\\
\noindent {\color{gray}``The manuscript needs to be revisited and thoroughly modified in order to be clearer to the reader.''}

Indeed, the article might be a little bit difficult to read as its structure and phrasing evolved several times.
We tried to address this in our review with focus on your detailed critique on concrete sections below.

~\\
\noindent {\color{gray}``In its current form, the scope of the work, its target, and the outcomes aren't clearly presented. Especially ?Section 5 Future Work? and ?Section 6 Conclusions? need to be revisited.''}

% Future work
Due to space constraints, we tried to reduce details in the section on future work.
In the revision, we gave it a more visual structure and separated between \textit{theoretical} and \textit{algorithmic} research directions.

% Conclusions, outcome & target
The conclusions summarizes the introduction of the method, a theoretical question we came across with, the methods' application to pruning and our interpretations based on the observations.
The section was modified for the revision to make several points of the article more clear.
Sentences with abstract mentions are now stated more explicitly, which should improve readability.

We revisited the mentioned sections and hope it fits the reviewers' point.

~\\
\noindent {\color{gray}``Page 3, Section 3.1, Equations 1 through 4: The Authors are asked to provide a detailed explanation of the meaning of the aforementioned equations. The way the equations are presented is not clear to the unfamiliar reader.''}

We made the equations more understandable by providing a concrete example as commonly used in Game Theory.
It should allow the reader to transfer the concept of a coalitional game to Artificial Neural Networks.

~\\
\noindent {\color{gray}``The 3rd and 4rt columns in Table 2 are identical. This should be a mistake. Please correct it.''}

Thank you for noticing this, we fixed it.
}{}

\end{document}